\documentclass{article}

\usepackage[final]{nips_2016}

\usepackage[utf8]{inputenc} 
\usepackage[T1]{fontenc}    
\usepackage{hyperref}       
\usepackage{url}            
\usepackage{booktabs}       
\usepackage{amsfonts}       
\usepackage{nicefrac}       
\usepackage{microtype}      
\usepackage{graphicx}
\usepackage{amsmath}
\usepackage{subcaption}
\usepackage{multirow}
\usepackage{wrapfig}

\title{Variational Graph Auto-Encoders}

\author{
  Thomas N.~Kipf \\
  University of Amsterdam \\
  \texttt{T.N.Kipf@uva.nl} \\
  \And
  Max Welling \\
  University of Amsterdam \\
  Canadian Institute for Advanced Research (CIFAR) \\
  \texttt{M.Welling@uva.nl} \\
}
\begin{document}

\maketitle
\vspace{-1em}
\section{A latent variable model for graph-structured data}

\begin{wrapfigure}{r}{0.5\textwidth}
\vspace{-1em}
\centering
  \includegraphics[width=0.5\textwidth, trim={0.4cm 0.7cm 0.7cm 1cm}, clip]{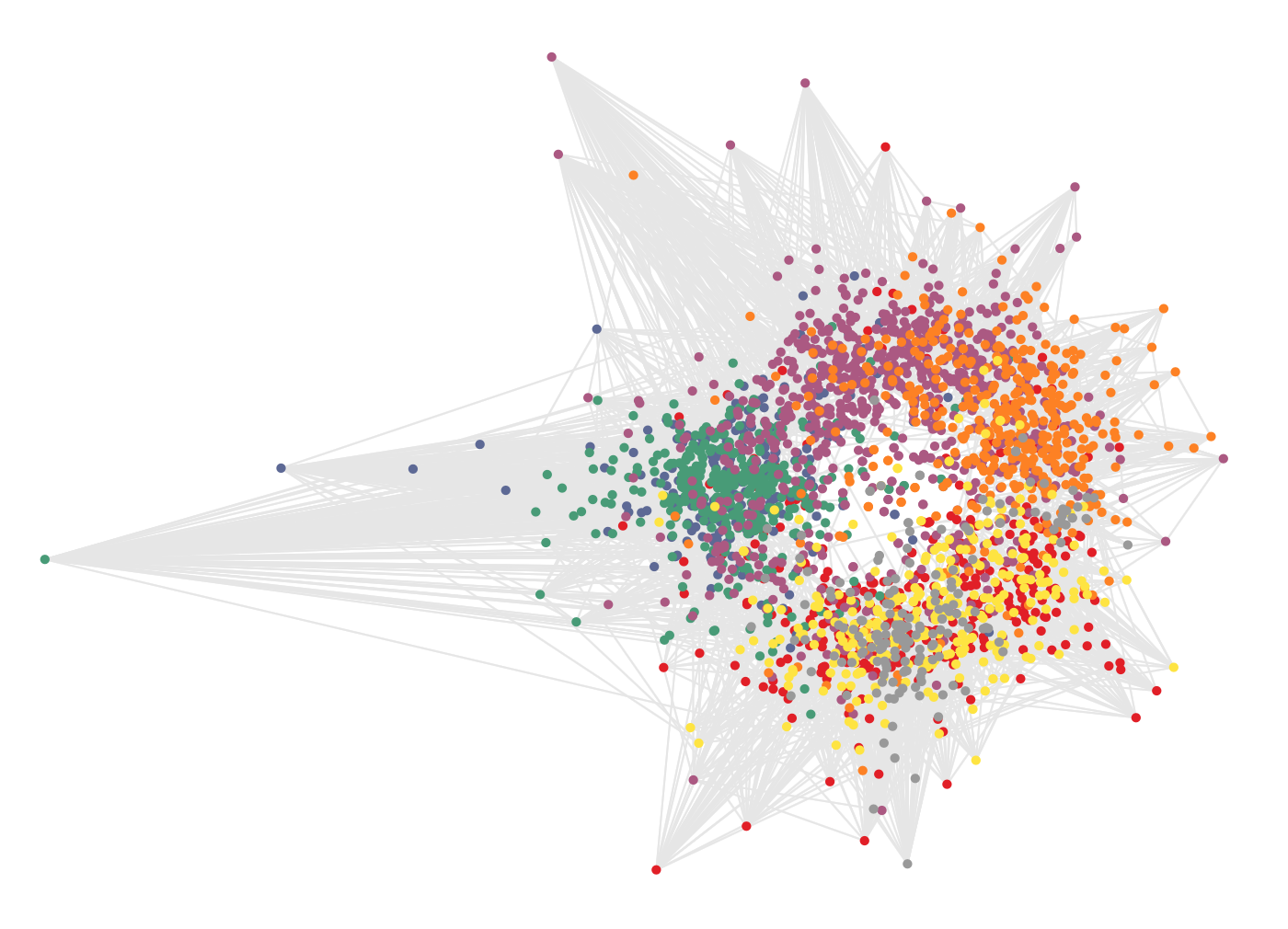}
\caption{Latent space of unsupervised VGAE model trained on Cora citation network dataset \cite{aimag08}. Grey lines denote citation links. Colors denote document class (not provided during training). Best viewed on screen.}
\label{fig:vgae}
\vspace{-1.5em}
\end{wrapfigure}

We introduce the \emph{variational graph auto-encoder} (VGAE), a framework for unsupervised learning on graph-structured data based on the \emph{variational auto-encoder} (VAE) \cite{kingma2013auto, rezende2014stochastic}. This model makes use of latent variables and is capable of learning interpretable latent representations for undirected graphs (see Figure \ref{fig:vgae}).

We demonstrate this model using a \emph{graph convolutional network} (GCN) \cite{kipf2016semi} encoder and a simple \emph{inner product} decoder. Our model achieves competitive results on a link prediction task in citation networks. In contrast to most existing models for unsupervised learning on graph-structured data and link prediction \cite{tang2011leveraging, perozzi2014deepwalk, tang2015line, grovernode2vec}, our model can naturally incorporate node features, which significantly improves predictive performance on a number of benchmark datasets.

\paragraph{Definitions}
We are given an undirected, unweighted graph $\mathcal{G}=(\mathcal{V},\mathcal{E})$ with $N=|\mathcal{V}|$ nodes. We introduce an adjacency matrix $\mathbf{A}$ of $\mathcal{G}$ (we assume diagonal elements set to 1, i.e.~every node is connected to itself) and its degree matrix $\mathbf{D}$. We further introduce stochastic latent variables $\mathbf{z}_i$, summarized in an $N\times F$ matrix $\mathbf{Z}$. Node features are summarized in an $N\times D$ matrix $\mathbf{X}$.
\paragraph{Inference model}
We take a simple inference model parameterized by a two-layer GCN:
\begin{equation}
\textstyle
q(\mathbf{Z}\,|\,\mathbf{X},\mathbf{A}) = \prod_{i=1}^N q(\mathbf{z}_i\,|\,\mathbf{X},\mathbf{A})\, , \,\,\, \text{with} \quad q(\mathbf{z}_i\,|\,\mathbf{X},\mathbf{A}) = \mathcal{N}(\mathbf{z}_i\,|\, \boldsymbol{\mu}_i, \mathrm{diag}(\boldsymbol{\sigma}_i^2)) \, .
\end{equation}
Here, $\boldsymbol{\mu} = \mathrm{GCN}_{\boldsymbol{\mu}}(\mathbf{X}, \mathbf{A})$ is the matrix of mean vectors $\boldsymbol{\mu}_i$; similarly $\log\boldsymbol{\sigma} = \mathrm{GCN}_{\boldsymbol{\sigma}}(\mathbf{X}, \mathbf{A})$. The two-layer GCN is defined as $\mathrm{GCN}(\mathbf{X}, \mathbf{A}) = \mathbf{\tilde{A}}\,\mathrm{ReLU}\bigl(\mathbf{\tilde{A}}\mathbf{X}\mathbf{W}_0\bigr)\mathbf{W}_1$, with weight matrices $\mathbf{W}_i$.  $\mathrm{GCN}_{\boldsymbol{\mu}}(\mathbf{X}, \mathbf{A})$ and $\mathrm{GCN}_{\boldsymbol{\sigma}}(\mathbf{X}, \mathbf{A})$ share first-layer parameters $\mathbf{W}_0$. $\mathrm{ReLU}(\cdot)=\max(0,\cdot)$ and $\mathbf{\tilde{A}}=\mathbf{D}^{-\frac{1}{2}}\mathbf{A}\mathbf{D}^{-\frac{1}{2}}$ is the symmetrically normalized adjacency matrix.

\paragraph{Generative model}
Our generative model is given by an inner product between latent variables:
\begin{equation}
\textstyle
p\left(\mathbf{A\,|\,\mathbf{Z}}\right) = \prod_{i=1}^N\prod_{j=1}^N p\left(A_{ij}\,|\,\mathbf{z}_i,\mathbf{z}_j\right)\, , \,\,\, \text{with} \quad p\left(A_{ij}=1\,|\,\mathbf{z}_i,\mathbf{z}_j\right) = \sigma(\mathbf{z}_i^\top\mathbf{z}_j) \, ,
\end{equation}
where $A_{ij}$ are the elements of $\mathbf{A}$ and $\sigma(\cdot)$ is the logistic sigmoid function.

\paragraph{Learning}
We optimize the variational lower bound $\mathcal{L}$ w.r.t.~the variational parameters $\mathbf{W}_i$:
\begin{equation}
\mathcal{L} =  \mathbb{E}_{q(\mathbf{Z}|\mathbf{X},\mathbf{A})}\bigl[\log p\left(\mathbf{A}\,|\,\mathbf{Z}\right)\bigr]
 - \mathrm{KL}\bigl[q(\mathbf{Z}\,|\,\mathbf{X},\mathbf{A})\,||\,p(\mathbf{Z})\bigr]\, ,
\end{equation}
where $\mathrm{KL}[q(\cdot)||p(\cdot)]$ is the Kullback-Leibler divergence between $q(\cdot)$ and $p(\cdot)$. We further take a Gaussian prior $p(\mathbf{Z}) = \prod_i p(\mathbf{z_i}) = \prod_i \mathcal{N}(\mathbf{z}_i\,|\, 0, \mathbf{I})$. For very sparse $\mathbf{A}$, it can be beneficial to re-weight terms with $A_{ij}=1$ in $\mathcal{L}$ or alternatively sub-sample terms with $A_{ij}=0$. We choose the former for the following experiments. We perform full-batch gradient descent and make use of the \emph{reparameterization trick} \cite{kingma2013auto} for training. For a featureless approach, we simply drop the dependence on $\mathbf{X}$ and replace $\mathbf{X}$ with the identity matrix in the GCN. 

\paragraph{Non-probabilistic \emph{graph auto-encoder} (GAE) model}
For a non-probabilistic variant of the VGAE model, we calculate embeddings $\mathbf{Z}$ and the reconstructed adjacency matrix $\mathbf{\hat{A}}$ as follows:
\begin{equation}
\mathbf{\hat{A}} = \sigma\bigl(\mathbf{Z} \mathbf{Z}^\top\bigr) \, , \,\,\,  \text{with} \quad \mathbf{Z} = \mathrm{GCN}(\mathbf{X}, \mathbf{A}) \, .
\label{eq:gae}
\end{equation}

\section{Experiments on link prediction}
We demonstrate the ability of the VGAE and GAE models to learn meaningful latent embeddings on a link prediction task on several popular citation network datastets \cite{aimag08}. The models are trained on an incomplete version of these datasets where parts of the citation links (edges) have been removed, while all node features are kept. We form validation and test sets from previously removed edges and the same number of randomly sampled pairs of unconnected nodes (non-edges).

We compare models based on their ability to correctly classify edges and non-edges. The validation and test sets contain $5\%$ and $10\%$ of citation links, respectively. The validation set is used for optimization of hyperparameters. We compare against two popular baselines: \emph{spectral clustering} (SC) \cite{tang2011leveraging} and \emph{DeepWalk} (DW) \cite{perozzi2014deepwalk}. Both SC and DW provide node embeddings $\mathbf{Z}$. We use Eq.~\ref{eq:gae} (left side) to calculate scores for elements of the reconstructed adjacency matrix. We omit recent variants of DW \cite{tang2015line, grovernode2vec} due to comparable performance. Both SC and DW do not support input features.

For VGAE and GAE, we initialize weights as described in \cite{glorot2010understanding}. We train for 200 iterations using Adam \cite{kingma2014adam} with a learning rate of $0.01$. We use a $32$-dim hidden layer and $16$-dim latent variables in all experiments. For SC, we use the implementation from \cite{scikit-learn} with an embedding dimension of $128$. For DW, we use the implementation provided by the authors of \cite{grovernode2vec} with standard settings used in their paper, i.e.~embedding dimension of $128$, $10$ random walks of length $80$ per node and a context size of $10$, trained for a single epoch.

\paragraph{Discussion}
Results for the link prediction task in citation networks are summarized in Table \ref{tab:results}. GAE* and VGAE* denote experiments without using input features, GAE and VGAE use input features. We report \emph{area under the ROC curve} (AUC) and \emph{average precision} (AP) scores for each model on the test set. Numbers show mean results and standard error for 10 runs with random initializations on fixed dataset splits.

\begin{table}[htp!]
\vspace{-0.5em}
  \caption{Link prediction task in citation networks. See \cite{aimag08} for dataset details.}
\vspace{0.5em}
  \label{tab:results}
  \centering
  \resizebox{\textwidth}{!}{
  \begin{tabular}{l r r r r r r}
    \toprule
     \multirow{2}{*}{\textbf{Method}}   & \multicolumn{2}{c}{\textbf{Cora}}     & \multicolumn{2}{c}{\textbf{Citeseer}} & \multicolumn{2}{c}{\textbf{Pubmed}} \\
     & \multicolumn{1}{c}{AUC} &  \multicolumn{1}{c}{AP} & \multicolumn{1}{c}{AUC} & \multicolumn{1}{c}{AP} & \multicolumn{1}{c}{AUC} & \multicolumn{1}{c}{AP} \\
    \midrule
    SC \cite{tang2011leveraging} & $84.6\pm0.01$ & $88.5\pm0.00$  & $80.5\pm0.01$ & $85.0\pm0.01$ & $84.2\pm0.02$ & $87.8\pm0.01$     \\
    DW \cite{perozzi2014deepwalk} & $83.1\pm0.01$ & $85.0\pm0.00$  & $80.5\pm0.02$ & $83.6\pm0.01$ & $84.4\pm0.00$ & $84.1\pm0.00$     \\
    \midrule
    GAE*     & $84.3\pm0.02$ & $88.1\pm0.01$  & $78.7\pm0.02$ & $84.1\pm0.02$ & $82.2\pm0.01$ & $87.4\pm0.00$     \\
    VGAE*     & $84.0\pm0.02$ & $87.7\pm0.01$  & $78.9\pm0.03$ & $84.1\pm0.02$ & $82.7\pm0.01$ & $87.5\pm0.01$     \\
    GAE    & $91.0\pm0.02$ & $92.0\pm0.03$  & $89.5\pm0.04$ & $89.9\pm0.05$ & $\mathbf{96.4}\pm0.00$ & $\mathbf{96.5}\pm0.00$     \\
    VGAE    & $\mathbf{91.4}\pm0.01$ & $\mathbf{92.6}\pm0.01$  & $\mathbf{90.8}\pm0.02$ & $\mathbf{92.0}\pm0.02$ & $94.4\pm0.02$ & $94.7\pm0.02$     \\
    \bottomrule
  \end{tabular}
  }
\end{table}

Both VGAE and GAE achieve competitive results on the featureless task. Adding input features significantly improves predictive performance across datasets. A Gaussian prior is potentially a poor choice in combination with an inner product decoder, as the latter tries to push embeddings away from the zero-center (see Figure \ref{fig:vgae}). Nevertheless, the VGAE model achieves higher predictive performance on both the Cora and the Citeseer dataset.

Future work will investigate better-suited prior distributions, more flexible generative models and the application of a stochastic gradient descent algorithm for improved scalability.

\subsubsection*{Acknowledgments}

We would like to thank Christos Louizos, Mart van Baalen, Taco Cohen, Dave Herman, Pramod Sinha and Abdul-Saboor Sheikh for insightful discussions. This project was funded by SAP Innovation Center Network.

\bibliographystyle{unsrt}
\bibpunct{[}{]}{,}{n}{}{;} 
\bibliography{references}

\begin{thebibliography}{10}

\bibitem{aimag08}
P.~Sen, G.~M. Namata, M.~Bilgic, L.~Getoor, B.~Gallagher, and T.~Eliassi-Rad.
\newblock Collective classification in network data.
\newblock {\em AI Magazine}, 29(3):93--106, 2008.

\bibitem{kingma2013auto}
D.~P. Kingma and M.~Welling.
\newblock Auto-encoding variational bayes.
\newblock In {\em Proceedings of the International Conference on Learning
  Representations (ICLR)}, 2014.

\bibitem{rezende2014stochastic}
D.~J. Rezende, S.~Mohamed, and D.~Wierstra.
\newblock Stochastic backpropagation and approximate inference in deep
  generative models.
\newblock In {\em Proceedings of The 31st International Conference on Machine
  Learning (ICML)}, 2014.

\bibitem{kipf2016semi}
T.~N. Kipf and M.~Welling.
\newblock Semi-supervised classification with graph convolutional networks.
\newblock {\em arXiv preprint arXiv:1609.02907}, 2016.

\bibitem{tang2011leveraging}
L.~Tang and H.~Liu.
\newblock Leveraging social media networks for classification.
\newblock {\em Data Mining and Knowledge Discovery}, 23(3):447--478, 2011.

\bibitem{perozzi2014deepwalk}
B.~Perozzi, R.~Al-Rfou, and S.~Skiena.
\newblock Deepwalk: Online learning of social representations.
\newblock In {\em Proceedings of the 20th ACM SIGKDD International Conference
  on Knowledge Discovery and Data Mining (KDD)}, pages 701--710. ACM, 2014.

\bibitem{tang2015line}
J.~Tang, M.~Qu, M.~Wang, M.~Zhang, J.~Yan, and Q.~Mei.
\newblock Line: Large-scale information network embedding.
\newblock In {\em Proceedings of the 24th International Conference on World
  Wide Web}, pages 1067--1077. ACM, 2015.

\bibitem{grovernode2vec}
A.~Grover and J.~Leskovec.
\newblock node2vec: Scalable feature learning for networks.
\newblock In {\em Proceedings of the 22nd ACM SIGKDD International Conference
  on Knowledge Discovery and Data Mining (KDD)}, 2016.

\bibitem{glorot2010understanding}
X.~Glorot and Y.~Bengio.
\newblock Understanding the difficulty of training deep feedforward neural
  networks.
\newblock In {\em Aistats}, volume~9, pages 249--256, 2010.

\bibitem{kingma2014adam}
D.~P. Kingma and J.~L. Ba.
\newblock Adam: A method for stochastic optimization.
\newblock In {\em Proceedings of the International Conference on Learning
  Representations (ICLR)}, 2015.

\bibitem{scikit-learn}
F.~Pedregosa, G.~Varoquaux, A.~Gramfort, V.~Michel, B.~Thirion, O.~Grisel,
  M.~Blondel, P.~Prettenhofer, R.~Weiss, V.~Dubourg, J.~Vanderplas, A.~Passos,
  D.~Cournapeau, M.~Brucher, M.~Perrot, and E.~Duchesnay.
\newblock Scikit-learn: Machine learning in {P}ython.
\newblock {\em Journal of Machine Learning Research}, 12:2825--2830, 2011.

\end{thebibliography}

\end{document}